%% file: main.tex
\documentclass[11pt, a4paper, logo, copyright]{googledeepmind}

\usepackage[authoryear, sort&compress, round]{natbib}
\usepackage[inkscapeformat=png]{svg}

\usepackage[most, breakable, skins]{tcolorbox}

\tcbuselibrary{skins}
\usepackage{lipsum}
\usepackage{tabularx}
\usepackage{afterpage}
\usepackage{booktabs}
\usepackage{subcaption}

\usepackage{makecell}
\usepackage{multirow}
\usepackage{multicol}
\usepackage{array}
\usepackage{epigraph}
\usepackage{float}
\usepackage{listings, listings-rust}
\usepackage{fontawesome5}
\usepackage{amssymb,graphicx}
\usepackage[dvipsnames]{xcolor}
\usepackage{hyperref}
\usepackage[capitalise, noabbrev]{cleveref}
\usepackage{longtable}
\usepackage{graphicx}
\usepackage{ragged2e}
\usepackage{pdflscape}
\usepackage{adjustbox}
\usepackage[section]{placeins}  
\usepackage{tikz}
\usetikzlibrary{shapes.geometric, arrows, positioning, fit}
\usepackage{tcolorbox} 
\usepackage{listings}
\usepackage{xspace}
\usepackage{caption} 
\usepackage{pifont}

\usepackage{titlesec}
\titlespacing*{\section}{0pt}{*1}{*0.5}
\titlespacing*{\subsection}{0pt}{*0.8}{*0.4}

\newcommand{\ra}[1]{\renewcommand{\arraystretch}{#1}}

\newcommand*{\colorboxed}{}
\def\colorboxed#1#{%
  \colorboxedAux{#1}%
}
\newcommand*{\colorboxedAux}[3]{%
  \begingroup
    \colorlet{cb@saved}{.}%
    \color#1{#2}%
    \boxed{%
      \color{cb@saved}%
      #3%
    }%
  \endgroup
}

\newcommand{\evalname}{\textsl{FACTS Grounding}\xspace}

\title{The \evalname Leaderboard: Benchmarking LLMs' Ability to  Ground Responses to Long-Form Input}

\correspondingauthor{facts-leaderboard@google.com}

\newcommand{\bGdm}{$\mathbin{\Diamond}$}
\newcommand{\bKaggle}{\ding{168}}
\newcommand{\bResearch}{\ding{171}}
\newcommand{\bCloud}{$\heartsuit$}

\author[*,\bResearch]{Alon Jacovi}
\author[*,\bKaggle]{Andrew Wang}
\author[*,\bGdm]{Chris Alberti}
\author[*,\bGdm]{Connie Tao}
\author[*,\bKaggle]{Jon Lipovetz}
\author[*,\bGdm]{Kate Olszewska}
\author[*,\bGdm]{Lukas Haas}
\author[*,\bCloud]{Michelle Liu}
\author[*,\bKaggle]{Nate Keating}
\author[\bResearch]{Adam Bloniarz}
\author[\bCloud]{Carl Saroufim}
\author[\bResearch]{Corey Fry}
\author[\bResearch]{Dror Marcus}
\author[\bResearch]{Doron Kukliansky}
\author[\bGdm]{Gaurav Singh Tomar}
\author[\bCloud]{James Swirhun}
\author[\bCloud]{Jinwei Xing}
\author[\bCloud]{Lily Wang}
\author[\bCloud]{Madhu Gurumurthy}
\author[\bKaggle]{Michael Aaron}
\author[\bResearch]{Moran Ambar}
\author[\bResearch]{Rachana Fellinger}
\author[\bCloud]{Rui Wang}
\author[\bCloud]{Zizhao Zhang}
\author[\bResearch]{Sasha Goldshtein}
\author[*,\bGdm]{Dipanjan Das}

\affil[*]{Equal Contribution}
\affil[\bGdm]{Google DeepMind}
\affil[\bResearch]{Google Research}
\affil[\bCloud]{Google Cloud}
\affil[\bKaggle]{Kaggle}

\begin{abstract}

We introduce \evalname, an online leaderboard and associated benchmark that evaluates language models' ability to generate text that is factually accurate with respect to given context in the user prompt. In our benchmark, each prompt includes a user request and a full document, with a maximum length of 32k tokens, requiring long-form responses. The long-form responses are required to be fully grounded in the provided context document while fulfilling the user request. Models are evaluated using automated judge models in two phases: (1) responses are disqualified if they do not fulfill the user request; (2) they are judged as accurate if the response is fully grounded in the provided document. The automated judge models were comprehensively evaluated against a held-out test-set to pick the best prompt template, and the final factuality score is an aggregate of multiple judge models to mitigate evaluation bias. The \evalname leaderboard will be actively maintained over time, and contains both public and private splits to allow for external participation while guarding the integrity of the leaderboard. It can be found at \url{https://www.kaggle.com/facts-leaderboard}.

\end{abstract}

\begin{document}

\maketitle

\input{intro_revised}

\input{data}

\input{metrics}

\input{results}

\input{conclusion}

\input{contributions}

\bibliography{main}

\clearpage

\appendix

\input{app_1}

\end{document}

%% file: intro_revised.tex
\section{Introduction}
\label{sec:intro_revised}

\textit{Factuality} is one of the most challenging aspects of Large Language Models (LLMs), referring to a model's ability to generate factually accurate responses in information-seeking scenarios. Commonly, this area of research can be divided into two distinct scenarios: (1) factuality with respect to \textit{given context}, such as a user request and grounding documents, such that the model response is fully grounded in the input (by this, we imply that a model response has the highest degree of faithfulness to given context as defined by  \citealp{Rashkin2021MeasuringAI}), and (2) factuality with respect to \textit{external sources} and \textit{general world knowledge}~(\citealp{tang-etal-2024-minicheck}, cf. \citealp{Rashkin2021MeasuringAI,Pan2023FactCheckingCC,Chen2023FELMBF}). There may be subtle use cases that fall in the middle, but most existing literature in the factuality area focus on behaviors that map to these two distinct scenarios. Summarization is a narrow but important example of the first scenario--a summary's various claims should be accurate with respect to the source documents that are being summarized~\citep{Bishop2023LongDocFACTScoreET,krishna-etal-2023-longeval}. Generating short-form, factually accurate answers to factoid questions from an LLM's parametric knowledge is an example use case of the second scenario, as recently discussed by~\cite{wei2024measuringshortformfactualitylarge}.

Ensuring factual accuracy while generating LLM responses is challenging. The principal challenges in LLM factuality are \textit{modeling} (i.e., architecture, training, and inference) and \textit{measurement} (i.e., evaluation methodology, data and metrics). On the modeling front, LLMs are inherently difficult to optimize for this goal: typically, LLM pretraining is optimized to predict the next token given previous tokens in the text. While this objective may teach models salient world knowledge, it does not directly optimize the model towards the various factuality scenarios, instead encouraging the model to generate generally \textit{plausible} text. Furthermore, critically, recent research suggests that the training process (regardless of specific models or training data) inherently admits non-factual text generation~\citep{kalai2024calibratedlanguagemodelshallucinate}.  Subsequent post-training, via supervised finetuning and reinforcement learning, can steer the model towards improved factuality~\citep{Lan2023FactGenFT,roit-etal-2023-factually,huang-chen-2024-factalign}. Other mitigation proposals include inference-time methods, such as prompting or model state interpretability~\citep{lee2024llmhallucinationreasoningzeroshot,Su2024UnsupervisedRH}. However, enhancing factuality through these methods can compromise other desirable attributes, such as creativity and novelty, which are also optimized during these stages. This creates a delicate balance, making it difficult to simultaneously achieve all desired outcomes~\citep{roit-etal-2023-factually}. 

Due to the above, factuality is expected to remain a research challenge for the foreseeable future. In this work, we focus on the challenge of measuring progress in factuality.  Measurement is in itself difficult, due to the different model behaviors that pertain to the aforementioned factuality scenarios. Some settings are more difficult than others: in \textit{long-form generation} tasks, the subject of our benchmark, each claim in the models' responses should be thoroughly inspected for their accuracy. This is in contrast to tasks that require measuring the factuality of short responses to questions against ground truth~\citep[e.g.,][]{wei2024measuringshortformfactualitylarge}.
To complicate matters further, long-form generation settings are in themselves varied and numerous. For example, some benchmarks focus on long-form generation in the setting of grounding to external sources~\citep[e.g.,][]{wei2024longform,zhu2024haluevalwildevaluatinghallucinationslanguage,zhao2024wildhallucinationsevaluatinglongformfactuality} while others focus on a particular task or domain, such as summarization to news or biomedical articles~\citep[e.g.,][]{hhem-2.1-open,ramprasad-etal-2024-evaluating}. 

Here, we investigate a benchmark that focuses on scenario (1)--particularly, the measurement of response factuality with respect to a provided context document of length up to 32k tokens, given varied user requests.  This setting requires the model to synthesize information derivable from the document while directly addressing the request. While encompassing summarization as a key use case, it extends to a broader range of requests, including fact finding, analyzing and comparing information, and so forth, while being fully grounded to the input document.
We believe that this benchmark fills a gap in evaluating a wider variety of model behaviors pertaining to factuality, in comparison to benchmarks that focus on narrower use cases, e.g. summarization alone~\citep{hhem-2.1-open}.  (Note however that we do not capture scenario (2) in this work.)

\begin{table*}[ht]\centering
\caption{Examples from \evalname (Public).}
\label{tab:examples}
\scriptsize
\ra{1.3}
\resizebox{0.99\linewidth}{!}{
\begin{tabular}{
p{5.1cm}
p{4.5cm}
c
p{5.0cm}
}
\toprule
\textbf{System Instruction} & \textbf{Context Document Description} & \textbf{Context Tokens} & \textbf{User Request} \\ \midrule
 Answer the question using only the information provided in the context. Do not rely on external knowledge or sources. &  The development and deployment of an autonomous robotic system designed to clean skyscraper windows, highlighting its technological advancements, safety implications, and potential impact on the window-washing industry. & $\sim$1.1k &  My sister and her dog live in NYC. I've visited there and have always been fascinated with their tall buildings. Then I thought...someone has to clean those! Then next thing you know, window washing robotos popped up on my feed. How do these robots work? Also what does this mean for the people who do those jobs? \\ \\
 Provide a response based solely on the information provided in the prompt. External sources and prior knowledge must not be used. &  Legal interpretations and effects of the medical marijuana appropriations rider on federal marijuana prosecutions, focusing on differing circuit court approaches to determining compliance with state medical marijuana laws. & $\sim$1.6k &  What did the first circuit conclude?  \\ \\
 This task requires you to answer questions based solely on the information provided in the prompt. You are not allowed to use any external resources or prior knowledge. Present your answer in headed sections with an explanation for each section. Each explanation should be in bullet points with exactly three bullet points. &  Comparison of different economic systems, including free market, command, and mixed economies, highlighting their key characteristics, advantages, and disadvantages. & $\sim$0.9k &  which famous economists are mentioned? \\ \\
 Provide your response in a professional and formal tone.
Use the information given in the document without referring to external sources or requiring additional context. Avoid using technical jargon or acronyms that are not explained within the document. &  Compilation of money-saving tips for college students, categorized into recreation and entertainment, food and basic needs, clothing, budgeting/spending plan, transportation, savings, and conserving resources. & $\sim$1.6k &  What are some tips on saving money? \\ \\
 Answer the question based solely on the information provided in the passage. Do not use any external knowledge or resources. &  A study that investigates the correlation between advanced maternal age (40+) and increased risk of obstetric, fetal, and neonatal complications compared to women aged 25-35. & $\sim$2.1k &  Researchers at Foch Hospital in France published this study of pregnancy outcomes in two groups of patients. Please summarize outcomes across the three kinds of complications that the researchers studied.
\\ \bottomrule
\end{tabular}
}
\end{table*}

Factuality of long-form responses is difficult to thoroughly measure at scale;  particularly automatic evaluation methods continue to be a challenge and is an active research area~\citep[\textit{inter alia}]{min-etal-2023-factscore,Bishop2023LongDocFACTScoreET,song-etal-2024-veriscore,gekhman-etal-2023-trueteacher,tang-etal-2024-minicheck,honovich-etal-2022-true,HaluEval,nocha-2024-karp-thai-et-al,jacovi2024coverbenchchallengingbenchmarkcomplex,fables-2024-kim-et-al,chang2024booookscore,ramprasad2024automaticfactualitymetricsmeasure}.  A particular limitation of existing automatic evaluation methods is that specialized factuality classifiers are limited to short context windows, or ones that cannot perform reasoning that is required to evaluate factuality behaviors~\citep{jacovi2024coverbenchchallengingbenchmarkcomplex}. Here, we rigorously evaluate our automatic evaluators on held-out test data to validate their performance on our task, and use multiple aggregations to mitigate evaluator bias.

We present the \evalname leaderboard and an associated benchmark measuring the ability of LLMs to ground long-form responses to document context up to length 32k tokens given a user request and additional instructions.
The benchmark contains 860 public examples (``Open'' split) and 859 private examples (``Blind'' split) of natural, complex LLM prompts written by humans to evaluate long-form grounded response generation (see examples in \Cref{tab:examples}; details in \Cref{sec:data}). The leaderboard reports various LLMs' performance on this benchmark using an automated factuality score incorporating an eligibility filter to avoid ``hacking'' the leaderboard metric (\Cref{sec:metrics}). The results are available in \Cref{sec:results}. The leaderboard will be actively maintained and updated to include new models and their variants.

%% file: data.tex
\section{Data}
\label{sec:data}

Underlying the \evalname leaderboard is a carefully curated collection of documents and associated user requests that were written by human raters and then subjected to thorough validation and filtering. The complete methodology is outlined in the following subsections. \Cref{tab:examples} provides concrete examples of data instances in the collection. To ensure the reliability of the benchmark, both public and private leaderboard splits were constructed using a balanced random sampling strategy.

\begin{figure}[ht]
    \centering
    \begin{minipage}{0.47 \textwidth}
        \centering
        \includegraphics[width=\linewidth]{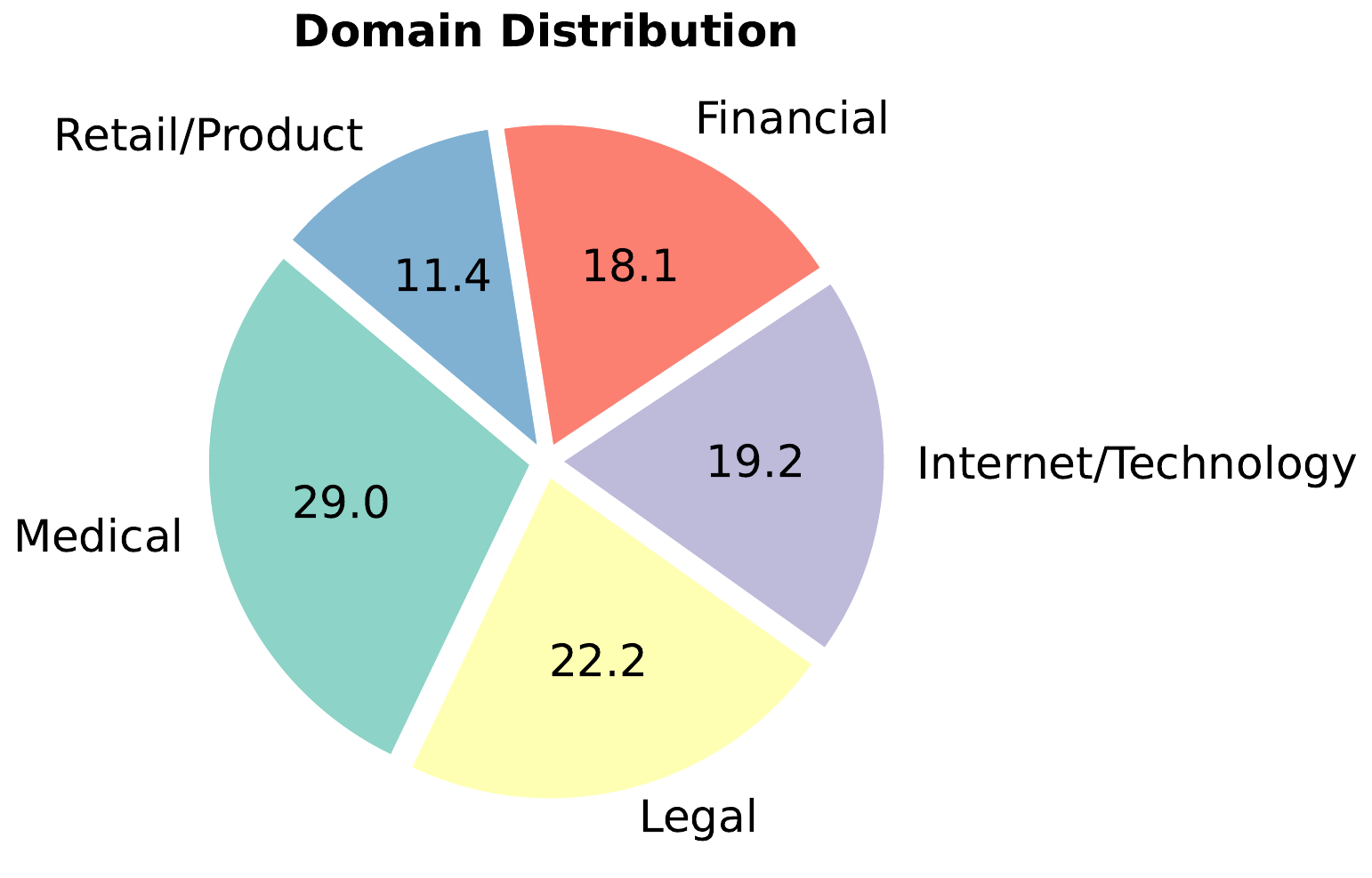}
    \end{minipage}
    \begin{minipage}{0.52\textwidth}
        \centering
        \includegraphics[width=\linewidth]{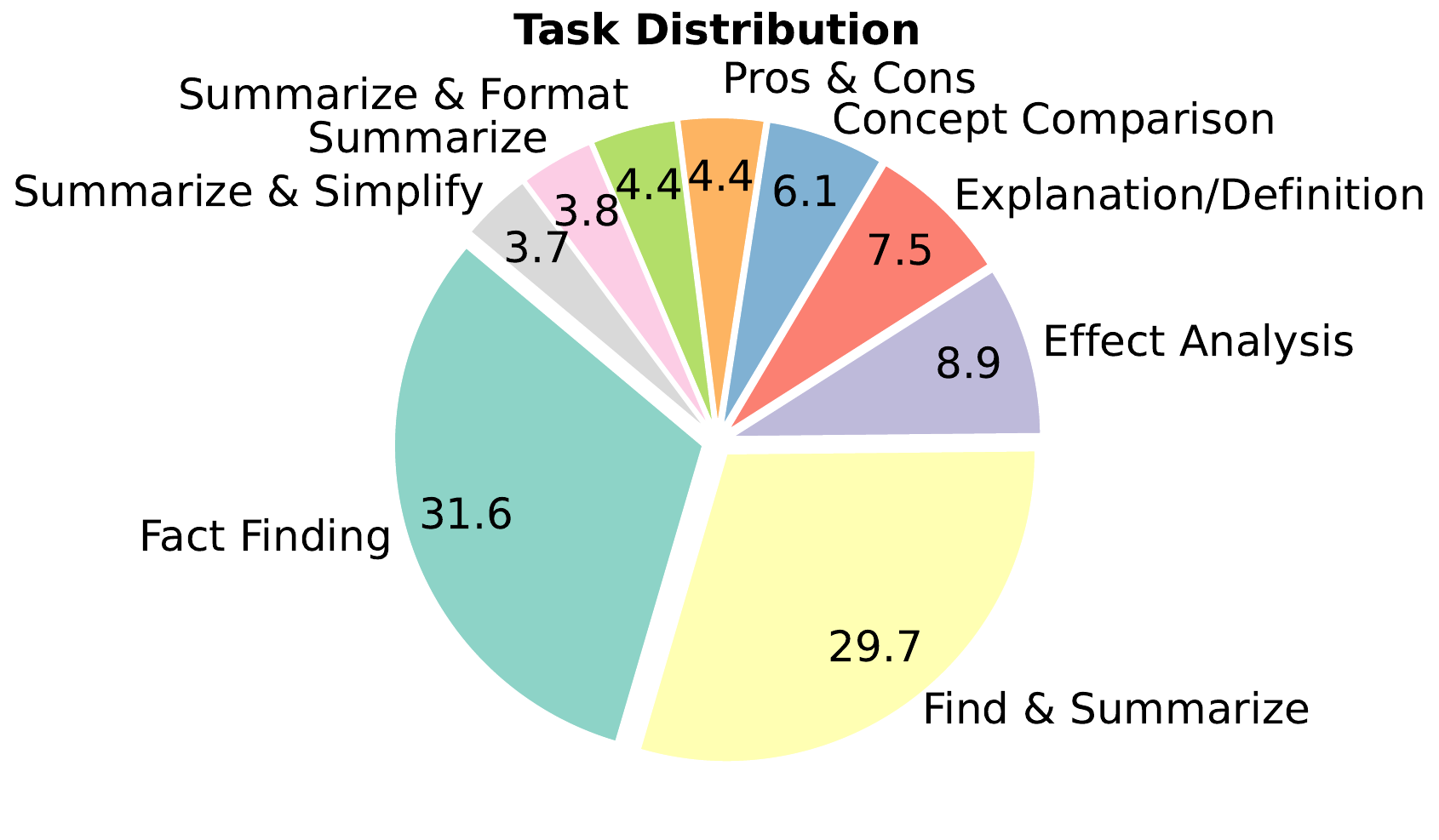}
    \end{minipage}
    \caption{Distributions of context domain and of task requested by the user as a percent of the total set of prompts in the benchmark.}
    \label{fig:distr}
\end{figure}

\subsection{Annotation}
In order to create our evaluation set, third-party human raters were instructed to design prompts requiring the processing of long-form input and the writing of long-form output. These tasks include Q\&A, summarization, and document rewriting. Each example within our evaluation set consists of a context, which is a document or set of reviews sourced from the web, paired with a non-trivial user request that can be addressed using the provided context, necessitating a long-form response. Additionally, each example includes a system instruction directing the model to generate its response exclusively from the given context, without incorporating external knowledge.

To ensure the diversity of the evaluation set, prompts were generated across a range of document lengths (up to 32k tokens) and various enterprise domains, including finance, technology, retail, medical, and legal. The annotation instructions were carefully designed to avoid prompts requiring creative responses, expert-level domain knowledge, mathematical or logical reasoning, or meta-analysis of the text, such as tone analysis or interpretation of author intent. The specific distributions of enterprise domains and of tasks requested by users are shown in Figure \ref{fig:distr}.

\subsection{Data Quality Assurance}

We manually verified all examples after annotation to discard those that did not align with our instructions. In particular, we ensured that all task instructions required the model to exclusively rely on the provided context, and we further removed creative-writing tasks. We also verified that the user requests were non-trivial and did not require domain expertise, mathematical knowledge, or complex reasoning. We additionally removed documents originating from PDFs where optical character recognition (OCR) rendered them unreadable. After our data quality assurance, the final dataset contained context documents with a mean length of 2.5k tokens and a maximum length of 32k tokens. 

\paragraph{On data contamination.} As the context documents were collected from the internet, it is possible that they were included in models' pre-training corpora. Although this is noteworthy, we make the following arguments towards the value of this benchmark:
\begin{enumerate}
    \item The user requests and system instructions, which instruct specifically to only follow information in the context document, are non-contaminated. Responding to novel requests about non-novel documents is an important use-case of language models, and grounding is integral to it. This is unlike many currently available factuality benchmarks which repurpose academic tasks that have likely been contaminated~\citep{sainz2024datacontaminationreport2024}.
    \item Our factuality score evaluates a distinct dimension of model performance that is not optimized during pre-training. Specifically, it measures the model's ability to generate responses grounded exclusively in the provided context. This means that the model must not incorporate external knowledge, even if conflicting with the context document, and should also avoid drawing upon any pre-training knowledge to fulfill the user's request.
    \item As all frontier language models models were trained on large corpora of web data, parity is maintained for the purpose of a leaderboard.
\end{enumerate}

%% file: metrics.tex
\section{Metrics}

\label{sec:metrics}

The final factuality score in \evalname is calculated as an aggregation of factuality verdicts from multiple judge models after the disqualification of ineligible responses that do not sufficiently succeed at responding to the user request.

\begin{table}[t]\centering
\caption{Evaluation of different judge models and evaluation prompts on a private test-set (N=402, class ratio 87:13; see \S\ref{sec:metrics-score}). Chosen prompt template for each model via Macro-F$_1$ in \textbf{bold}. }\label{tab:grounding-judge-eval}
\ra{1.2}
\scriptsize
\begin{tabular}{llrrrrrr}\toprule
\textbf{Judge Model} &\textbf{Prompt Template} & \textbf{Macro-F$_1$} &\textbf{Acc.} &\textbf{FPR} &\textbf{FNR} &\textbf{F$_1$ ($+$)} &\textbf{F$_1$ ($-$)}  \\\midrule
\multirow{7}{*}{Claude 3.5 Sonnet} &Span-level &68.85 &77.83 &20.97 &22.38 &85.58 &52.13 \\
 &\textbf{Implicit span-level} &\textbf{70.24} &83.50 &45.16 &11.34 &90.10 &50.37 \\
 &Response-level &61.88 &83.25 &72.58 &6.69 &90.42 &33.33 \\
 &JSON &56.04 &64.78 &33.87 &35.47 &75.64 &36.44 \\
 &JSON (alt) &55.37 &66.75 &46.77 &30.81 &77.91 &32.84 \\ 
 &JSON w. double-check &49.50 &54.68 &25.81 &48.84 &65.67 &33.33 \\
 &SimpleQA template &55.39 &85.22 &88.71 &1.45 &91.87 &18.92 \\\midrule
\multirow{7}{*}{Gemini 1.5 Pro} &Span-level &55.84 &79.31 &79.03 &10.17 &88.03 &23.64 \\
 &Implicit span-level &56.66 &85.47 &87.10 &1.45 &91.99 &21.33 \\
 &Response-level &48.82 &82.02 &95.16 &4.07 &90.04 &7.59 \\
 &\textbf{JSON} &\textbf{71.47} &86.95 &56.45 &5.23 &92.48 &50.47 \\ 
 &JSON (alt) &66.03 &85.96 &69.35 &4.07 &92.05 &40.00 \\
 &JSON w. double-check &64.89 &76.35 &37.10 &21.22 &84.95 &44.83 \\
 &SimpleQA template &51.54 &84.73 &93.55 &1.16 &91.64 &11.43 \\\midrule
\multirow{7}{*}{GPT-4o} &Span-level &63.08 &81.53 &64.52 &10.17 &89.18 &36.97 \\
 &Implicit span-level &55.43 &83.99 &87.10 &3.20 &91.11 &19.75 \\
 &Response-level &51.54 &84.73 &93.55 &1.16 &91.64 &11.43 \\
 &\textbf{JSON} &\textbf{69.68} &80.54 &32.26 &17.15 &87.83 &51.53 \\
 &JSON (alt) &66.78 &82.02 &53.23 &11.63 &89.28 &44.27 \\
 &JSON w. double-check &57.62 &64.04 &17.74 &39.24 &74.11 &41.13 \\
&SimpleQA template &47.04 &83.74 &98.39 &1.45 &91.13 &2.94 \\
\bottomrule
\end{tabular}
\end{table}

\subsection{Unadjusted Factuality Score}

\label{sec:metrics-score}

The principal component of our evaluation process is an unadjusted \textit{factuality score}. 

First, we utilize a language model judge to produce a binary classification label identifying whether a full model response is grounded in the user request and the context document given an instruction (see Table~\ref{tab:examples}). A model response is marked with a positive label (``accurate'') if all the claims in the response are grounded in the contents of the prompt, or do not require grounding;  the response is marked with a negative label (``not accurate'') if a single claim that bears information is deemed to be not grounded in the contents of the prompt. We use three different judge models in order to reduce the bias of a particular judge model, as models have been shown to be biased towards favorably judging their own outputs~\citep{wataoka2024selfpreferencebiasllmasajudge}. The judges are prompted language models in all cases---\textit{Gemini 1.5 Pro}~\citep{team2023gemini}, \textit{GPT-4o}~\citep{openai2024gpt4}, and \textit{Claude 3.5 Sonnet}~\citep{anthropic_claude_2024}. To select the right judge prompt, we tested seven different prompt templates and evaluated them based on alignment with human judgements on a held-out private set of prompts and model responses (N=402) which were annotated with golden labels. The prompt templates used in the evaluation are in~\cref{app:measurement}, and results are shown in~\cref{tab:grounding-judge-eval}. For each judge model, we select the best-performing prompt template via Macro-F$_1$.

Given the three judges, the individual factuality score for each judge is the percentage of accurate responses, and the unadjusted factuality score is the average of all judge models' scores.

\subsection{Disqualifying Ineligible Responses}
\label{sec:metrics-filter}

\begin{table*}[t]\centering
\caption{Examples of ineligible responses: these responses, while fully grounded in the context document, fail to address the user request meaningfully and are consequently considered ineligible.}
\label{tab:examples-quality}
\scriptsize
\ra{1.3}
\resizebox{0.99\linewidth}{!}{
\begin{tabular}{
p{4cm}
p{2.7cm}
p{2.7cm}
p{4.4cm}
}
\toprule
\textbf{Context Document Description} & \textbf{User Request} & \textbf{Ineligible Response} & \textbf{Rationale} \\ \midrule
 A research report on renewable energy sources, including wind, solar, and hydroelectric power, with specific statistics and case studies. &  Can you summarize the key advantages and disadvantages of wind energy from this document? &  Wind energy is good because it’s renewable and clean, but it has some challenges too. &  (1) The response is extremely vague, failing to provide any detailed or specific points from the document, such as the cost-effectiveness, geographic limitations, or impacts on wildlife. (2) It doesn’t engage with the query’s focus on "key" advantages and disadvantages. \\ \\
  A company’s annual financial report, discussing quarterly earnings, expenditures, future investments, and an analysis of the market environment. &  Summarize the main reasons the company’s revenue decreased in Q3. &  The company faced challenges in Q3 that impacted its revenue. &  (1) The response avoids specifying any reasons, such as market trends, increased competition, or operational setbacks, which would likely be in the document. (2) It doesn’t demonstrate an attempt to engage with or extract relevant details. \\ \\
 A historical article on the causes and consequences of the Great Depression. &  What were the main causes of the Great Depression as explained in the document? &  The Great Depression was a difficult time in history with many causes and effects. &   (1) The response provides no substantive information on the causes, such as stock market speculation, bank failures, or trade policies, which were discussed in the document. (2) It ignores the user's explicit focus on the "main causes."
\\ \bottomrule
\end{tabular}
}
\end{table*}

\begin{table}[t]\centering
\caption{Evaluation of prompt template for the eligible responses detection task on a private test-set (N=450; see \S\ref{sec:metrics-filter}). Chosen prompt template for each model via Macro-F$_1$ in \textbf{bold}.}\label{tab:quality-judge-eval}
\ra{1.2}
\scriptsize
\begin{tabular}{llrrrrrr}\toprule
\textbf{Judge Model} &\textbf{Prompt Template} & \textbf{Macro-F$_1$} &\textbf{Acc.} &\textbf{FPR} &\textbf{FNR} &\textbf{F$_1$ ($+$)} &\textbf{F$_1$ ($-$)}  \\\midrule
\multirow{2}{*}{Claude 3.5 Sonnet}
 &\textbf{User request only} &\textbf{60.88} &68.22	&16.33	&62.67	&43.92	&77.83 \\
 &User request and context document &58.14	&69.56	&8.67	&74.00	&36.28	&80.00 \\
\midrule
\multirow{2}{*}{Gemini 1.5 Pro}
 &\textbf{User request only} &\textbf{56.28}	&67.11	&12.33	&74.00	&34.51	&78.04 \\
 &User request and context document &47.56	&68.44	&1.33	&92.00	&14.46	&80.65 \\
 \midrule
\multirow{2}{*}{GPT-4o}
 &\textbf{User request only} &\textbf{55.16}	&69.56	&5.33	&80.67	&29.74	&80.57 \\
 &User request and context document &50.55	&69.56	&1.33	&88.67	&19.88	&81.21 \\
\bottomrule
\end{tabular}
\end{table}

Metrics that are focused on evaluating the factuality of the generated text, with respect to a context document or otherwise, can be circumvented by ignoring the intent behind the user request. By giving shorter responses that evade conveying comprehensive information, even if such content was an important aspect of a user request, it is possible to achieve a high factuality score while not providing a helpful response. See illustrative examples in \Cref{tab:examples-quality}.

We safeguard against such responses by detecting them with prompted LLM judges that use the same base models as in \Cref{sec:metrics-score}, with prompt templates described in \Cref{app:measurement}. At a response level, similar to unadjusted factuality score, we treat instruction-following as a distinct task for an LLM judge. This task involves a binary classification of each model response, determining whether it sufficiently addresses the user's request. Each response is assigned a binary label indicating its eligibility: either ``eligible,'' signifying that it answers the user's request, or ``ineligible,'' otherwise. Ineligible responses are disqualified from factuality evaluation and the final factuality score is adjusted such that ineligible responses are deemed as \textit{inaccurate}.

We investigate prompting each LLM judge with a prompt that contains either the user request only or the user request and the context document. We evaluate these two prompt templates on a separate test set which includes golden labels for instruction-following eligibility and assess whether the predicted ratings align with these golden labels (see Table~\ref{tab:quality-judge-eval} for results). For each LLM judge, we select the prompt with the highest Macro-F$_1$. Per response, eligibility classifications across the three judges are ensembled by consensus. A response is considered ineligible only if all three judges consider the response ineligible. Ensembling via consensus focuses this benchmark on evaluating grounding while still filtering out the worst quality responses.

%% file: results.tex
\begin{table}[t]\centering
\caption{\evalname results representing unadjusted \textit{factuality score} (no disqualification). Results are reported with a 95\% confidence interval.}\label{tab:grounding-results}
\scriptsize
\ra{1.2}

\resizebox{0.99\linewidth}{!}{
\begin{tabular}{lccc@{\hskip 20pt}ccccc}\toprule
\multirow{4}{*}{\textbf{Response Model}} &\multicolumn{6}{c}{\textbf{Judge Models}} &\multirow{4}{*}{\textbf{Average}} &\multirow{3}{*}{\textbf{Fused}}\\
&\multicolumn{3}{c}{\textbf{Open (N=860)}} &\multicolumn{3}{c}{\textbf{Blind (N=859)}} & & \multirow{3}{*}{\textbf{Rank}} \\\cmidrule{2-7}
&\textit{Gemini 1.5}&\textit{GPT-4o} &\textit{Claude 3.5} &\textit{Gemini 1.5} &\textit{GPT-4o} &\textit{Claude 3.5} & \\
&\textit{Pro}       &       &\textit{Sonnet}     &\textit{Pro}        &       & \textit{Sonnet}    & \\\midrule
\textbf{Gemini 1.5 Flash} & 91.4$_{(\pm 1.9)}$ & 82.0$_{(\pm 2.6)}$ & 84.9$_{(\pm 2.4)}$ & 90.7$_{(\pm 1.9)}$ & 80.7$_{(\pm 2.6)}$ & 85.1$_{(\pm 2.4)}$ & \textbf{85.8$_{(\pm 1.7)}$}& 1 \\
Gemini 2.0 Flash Experimental & 88.7$_{(\pm 2.1)}$ & 79.5$_{(\pm 2.7)}$ & 86.7$_{(\pm 2.3)}$ & 91.5$_{(\pm 1.9)}$ & 79.3$_{(\pm 2.7)}$ & 88.0$_{(\pm 2.2)}$ & 85.6$_{(\pm 1.7)}$& 2 \\
Gemini 1.5 Pro & 90.2$_{(\pm 2.0)}$ & 76.0$_{(\pm 2.9)}$ & 83.6$_{(\pm 2.5)}$ & 89.8$_{(\pm 2.0)}$ & 73.2$_{(\pm 3.0)}$ & 83.5$_{(\pm 2.5)}$ & 82.7$_{(\pm 1.8)}$& 3 \\
Claude 3.5 Sonnet & 88.4$_{(\pm 2.1)}$ & 72.7$_{(\pm 3.0)}$ & 87.7$_{(\pm 2.2)}$ & 88.4$_{(\pm 2.1)}$ & 73.8$_{(\pm 2.9)}$ & 82.2$_{(\pm 2.6)}$ & 82.2$_{(\pm 1.8)}$& 4 \\
GPT-4o & 86.7$_{(\pm 2.3)}$ & 71.7$_{(\pm 3.0)}$ & 79.2$_{(\pm 2.7)}$ & 88.0$_{(\pm 2.2)}$ & 75.9$_{(\pm 2.9)}$ & 77.4$_{(\pm 2.8)}$ & 79.8$_{(\pm 1.9)}$& 5 \\
Claude 3.5 Haiku & 85.8$_{(\pm 2.3)}$ & 67.2$_{(\pm 3.1)}$ & 82.0$_{(\pm 2.6)}$ & 80.1$_{(\pm 2.7)}$ & 62.5$_{(\pm 3.2)}$ & 74.3$_{(\pm 2.9)}$ & 75.3$_{(\pm 2.0)}$& 6 \\
GPT-4o mini & 80.8$_{(\pm 2.6)}$ & 62.1$_{(\pm 3.2)}$ & 71.4$_{(\pm 3.0)}$ & 83.0$_{(\pm 2.5)}$ & 65.0$_{(\pm 3.2)}$ & 70.7$_{(\pm 3.0)}$ & 72.2$_{(\pm 2.1)}$& 7 \\
OpenAI o1-mini & 70.8$_{(\pm 3.0)}$ & 50.2$_{(\pm 3.3)}$ & 63.1$_{(\pm 3.2)}$ & 75.1$_{(\pm 2.9)}$ & 51.3$_{(\pm 3.3)}$ & 64.6$_{(\pm 3.2)}$ & 62.5$_{(\pm 2.3)}$& 8 \\
OpenAI o1-preview & 69.2$_{(\pm 3.1)}$ & 50.1$_{(\pm 3.3)}$ & 65.3$_{(\pm 3.2)}$ & 70.0$_{(\pm 3.1)}$ & 53.2$_{(\pm 3.3)}$ & 65.0$_{(\pm 3.2)}$ & 62.1$_{(\pm 2.3)}$& 9 \\
\bottomrule
\end{tabular}
}
\end{table}

\section{Results}
\label{sec:results}

\Cref{tab:grounding-results} and \Cref{tab:grounding-quality-results} contain the unadjusted and final \textit{factuality scores} before and after disqualifying ineligible responses, respectively. The tested models are \textit{Gemini 1.5 Pro} and \textit{Flash}~\citep{team2023gemini}, \textit{Gemini 2.0 Flash Experimental}~\citep{gemini2blogpost},
\textit{GPT-4o}~\citep{openai2024gpt4}, 
\textit{OpenAI o1-preview} and \textit{o1-mini}~\citep{openai_learning_to_reason},
\textit{Claude 3.5 Haiku} and \textit{Sonnet}~\citep{anthropic_claude_2024}.
For the "Fused Rank" metric, we employ a ranking aggregation method that combines the six individual model rankings—derived from each split and judge model—into a single, unified ranking using the Condorcet algorithm. The resulting fused rank exactly aligns with the ranking obtained using the final factuality score.

\paragraph{On aggregating multiple judge models.} Consistent with prior research~\citep{liu-etal-2024-llms-narcissistic,NEURIPS2023_91f18a12,ye2024justiceprejudicequantifyingbiases,wataoka2024selfpreferencebiasllmasajudge,xu-etal-2024-pride}, we found that models generally rate their own outputs higher than those of other models, exhibiting a mean increase of +3.23\%. While the use of multiple judge models increases the computational cost of evaluation, this approach is essential when the judge models themselves are also subject to evaluation, as is the case in our work.

\paragraph{On ineligible responses.} Disqualifying ineligible responses leads to a reduction of $1\%$--$5\%$ in the final factuality score, as these responses are treated as inaccurate. The disqualification process also induces a minor shift in model rankings; for instance, Gemini 1.5 Flash moves from rank 1 to rank 2.

\begin{table}[t]\centering
\caption{\evalname Results representing the final \textit{factuality score} after disqualifying ineligible responses (see \S\ref{sec:metrics-filter}). Results are reported with a 95\% confidence interval. }\label{tab:grounding-quality-results}
\scriptsize
\ra{1.2}
\resizebox{0.99\linewidth}{!}{
\begin{tabular}{lccc@{\hskip 20pt}ccccc}\toprule
\multirow{4}{*}{\textbf{Response Model}} &\multicolumn{6}{c}{\textbf{Judge Models}} &\multirow{4}{*}{\textbf{Average}} &\multirow{3}{*}{\textbf{Fused}} \\
&\multicolumn{3}{c}{\textbf{Open (N=860)}} &\multicolumn{3}{c}{\textbf{Blind (N=859)}} &  &\multirow{3}{*}{\textbf{Rank}}\\\cmidrule{2-7}
&\textit{Gemini 1.5}&\textit{GPT-4o} &\textit{Claude 3.5} &\textit{Gemini 1.5} &\textit{GPT-4o} &\textit{Claude 3.5} &\\
&\textit{Pro}       &       &\textit{Sonnet}     &\textit{Pro}        &       & \textit{Sonnet}    & \\\midrule
\textbf{Gemini 2.0 Flash Experimental} & 86.4$_{(\pm 2.3)}$ & 77.4$_{(\pm 2.8)}$ & 84.8$_{(\pm 2.4)}$ & 89.3$_{(\pm 2.1)}$ & 77.2$_{(\pm 2.8)}$ & 86.3$_{(\pm 2.3)}$ & \textbf{83.6$_{(\pm 1.8)}$}& 1 \\
Gemini 1.5 Flash & 88.1$_{(\pm 2.2)}$ & 79.2$_{(\pm 2.7)}$ & 82.6$_{(\pm 2.5)}$ & 87.3$_{(\pm 2.2)}$ & 77.9$_{(\pm 2.8)}$ & 82.3$_{(\pm 2.6)}$ & 82.9$_{(\pm 1.8)}$& 2 \\
Gemini 1.5 Pro & 87.4$_{(\pm 2.2)}$ & 73.8$_{(\pm 2.9)}$ & 81.4$_{(\pm 2.6)}$ & 86.5$_{(\pm 2.3)}$ & 70.2$_{(\pm 3.1)}$ & 80.8$_{(\pm 2.6)}$ & 80.0$_{(\pm 1.9)}$& 3 \\
Claude 3.5 Sonnet & 87.3$_{(\pm 2.2)}$ & 71.7$_{(\pm 3.0)}$ & 86.7$_{(\pm 2.3)}$ & 82.0$_{(\pm 2.6)}$ & 67.8$_{(\pm 3.1)}$ & 81.0$_{(\pm 2.6)}$ & 79.4$_{(\pm 1.9)}$& 4 \\
GPT-4o & 85.3$_{(\pm 2.4)}$ & 70.7$_{(\pm 3.0)}$ & 78.5$_{(\pm 2.7)}$ & 86.5$_{(\pm 2.3)}$ & 74.9$_{(\pm 2.9)}$ & 76.8$_{(\pm 2.8)}$ & 78.8$_{(\pm 1.9)}$& 5 \\
Claude 3.5 Haiku & 84.8$_{(\pm 2.4)}$ & 66.6$_{(\pm 3.2)}$ & 81.2$_{(\pm 2.6)}$ & 77.9$_{(\pm 2.8)}$ & 61.0$_{(\pm 3.3)}$ & 73.7$_{(\pm 2.9)}$ & 74.2$_{(\pm 2.1)}$& 6 \\
GPT-4o mini & 79.4$_{(\pm 2.7)}$ & 60.8$_{(\pm 3.3)}$ & 70.7$_{(\pm 3.0)}$ & 81.4$_{(\pm 2.6)}$ & 63.7$_{(\pm 3.2)}$ & 70.1$_{(\pm 3.1)}$ & 71.0$_{(\pm 2.1)}$& 7 \\
OpenAI o1-mini & 70.2$_{(\pm 3.1)}$ & 49.8$_{(\pm 3.3)}$ & 62.7$_{(\pm 3.2)}$ & 74.2$_{(\pm 2.9)}$ & 50.8$_{(\pm 3.3)}$ & 64.3$_{(\pm 3.2)}$ & 62.0$_{(\pm 2.3)}$& 8 \\
OpenAI o1-preview & 68.6$_{(\pm 3.1)}$ & 49.7$_{(\pm 3.3)}$ & 65.0$_{(\pm 3.2)}$ & 69.4$_{(\pm 3.1)}$ & 52.6$_{(\pm 3.3)}$ & 64.6$_{(\pm 3.2)}$ & 61.7$_{(\pm 2.3)}$& 9 \\

\bottomrule
\end{tabular}
}
\end{table}

%% file: conclusion.tex
\section{Conclusion}
\label{sec:conclusion}

The \evalname leaderboard is designed to rigorously challenge language models' ability to maintain factual accuracy when generating long-form responses grounded in a document provided within the prompt, and in accordance with a user's specific request and instructions. We encourage other researchers to utilize this benchmark to advance both the factual capabilities of models and the methodologies for evaluating factuality.

%% file: contributions.tex
\section{Contributions and Acknowledgements}
\label{sec:contributions}

\vspace{0.2cm}

\begin{itemize}
\item \textbf{Experimental design:} Alon Jacovi, Andrew Wang, Chris Alberti, Jon Lipovetz, and Michelle Liu created the experimental design behind the benchmark and ran all the reported experiments.\vspace{0.2cm}
\item \textbf{Organization:} Connie Tao, Dipanjan Das, Kate Olszewska, Lukas Haas, and Nate Keating managed the overall organization of the effort from start to completion.\vspace{0.2cm}
\item \textbf{Early experimentation:} Adam Bloniarz, Carl Saroufim, Corey Fry, Dror Marcus, Doron Kukliansky, Gaurav Singh Tomar, James Swirhun, Jinwei Xing, Lily Wang, Madhu Gurumurthy, Michael Aaron, Moran Ambar, Rachana Fellinger, Rui Wang, Zizhao Zhang and Sasha Goldshtein contributed to ideas, conducted data collection and ran early experiments.\vspace{0.2cm}
\item \textbf{Sponsors:} Avinatan Hassidim, D. Sculley, Fernando Pereira, Koray Kavukcuoglu, Slav Petrov, Ya Xu, and Yossi Matias sponsored the effort and provided technical guidance.\vspace{0.2cm}
\item Slav Petrov and Madhu Gurumurthy proposed the idea for this leaderboard.
\end{itemize}

All authors wrote parts of the report.\vspace{0.2cm}

We would also like to thank:
\begin{itemize}
    \item \textbf{Gemini team} for the support and model access.
    \item \textbf{Kaggle team} for their expertise and releasing the leaderboard.
    \item \textbf{Expert data annotators} who helped to collect examples in the paper.
    \item Our reviewers \textbf{Kellie Webster} and \textbf{Phoebe Kirk} for valuable feedback.
\end{itemize}

%% file: app_1.tex
\section{Judge Prompt Templates} \label{app:measurement}

\noindent\textbf{Factuality Score.}

\noindent\textit{Response-level} 

\begin{lstlisting}
Your task is to check if the Response is accurate to the Evidence.
Generate 'Accurate' if the Response is accurate when verified according to the Evidence, or 'Inaccurate' if the Response is inaccurate (contradicts the evidence) or cannot be verified.

**Query**:\n\n{user_query}\n\n**End of Query**\n
**Evidence**\n\n{context}\n\n**End of Evidence**\n
**Response**:\n\n{response}\n\n**End of Response**\n
Let's think step-by-step.
\end{lstlisting}

\noindent\textit{JSON (Alt)} 

\begin{lstlisting}
You are a helpful and harmless AI assistant. You will be provided with a textual context and a model-generated response.
Your task is to analyze the response sentence by sentence and classify each sentence according to its relationship with the provided context.

**Instructions:**

1. **Decompose the response into individual sentences.**
2. **For each sentence, assign one of the following labels:**
    * **`supported`**: The sentence is entailed by the given context.  Provide a supporting excerpt from the context.
    * **`unsupported`**: The sentence is not entailed by the given context. Provide an excerpt that is close but does not fully support the sentence.
    * **`contradictory`**: The sentence is falsified by the given context. Provide a contradicting excerpt from the context.
    * **`no_rad`**: The sentence does not require factual attribution (e.g., opinions, greetings, questions, disclaimers).  No excerpt is needed for this label.

3. **For each label, provide a short rationale explaining your decision.**  The rationale should be separate from the excerpt.

**Input Format:**

The input will consist of two parts, clearly separated:

* **Context:**  The textual context used to generate the response.
* **Response:** The model-generated response to be analyzed.

**Output Format:**

For each sentence in the response, output a JSON object with the following fields:

* `"sentence"`: The sentence being analyzed.
* `"label"`: One of `supported`, `unsupported`, `contradictory`, or `no_rad`.
* `"rationale"`: A brief explanation for the assigned label.
* `"excerpt"`:  A relevant excerpt from the context. Only required for `supported`, `unsupported`, and `contradictory` labels.

Output each JSON object on a new line.

**Example:**

**Input:**
\end{lstlisting}
\begin{lstlisting}
```
Context: Apples are red fruits. Bananas are yellow fruits.

Response: Apples are red. Bananas are green.  Enjoy your fruit!
```

**Output:**

{"sentence": "Apples are red.", "label": "supported", "rationale": "The context explicitly states that apples are red.", "excerpt": "Apples are red fruits."}
{"sentence": "Bananas are green.", "label": "contradictory", "rationale": "The context states that bananas are yellow, not green.", "excerpt": "Bananas are yellow fruits."}
{"sentence": "Enjoy your fruit!", "label": "no_rad", "rationale": "This is a general expression and does not require factual attribution.", "excerpt": null}

**Now, please analyze the following context and response:**

**User Query:**
{user_query}

**Context:**
{context}

**Response:**
{response}
\end{lstlisting}

\noindent\textit{JSON} 

\begin{lstlisting}
You are a helpful and harmless AI assistant. You will be provided with a textual context and a model-generated response.
Your task is to analyze the response sentence by sentence and classify each sentence according to its relationship with the provided context.

**Instructions:**

1. **Decompose the response into individual sentences.**
2. **For each sentence, assign one of the following labels:**
    * **`supported`**: The sentence is entailed by the given context.  Provide a supporting excerpt from the context. The supporting except must *fully* entail the sentence. If you need to cite multiple supporting excepts, simply concatenate them.
    * **`unsupported`**: The sentence is not entailed by the given context. No excerpt is needed for this label.
    * **`contradictory`**: The sentence is falsified by the given context. Provide a contradicting excerpt from the context.
    * **`no_rad`**: The sentence does not require factual attribution (e.g., opinions, greetings, questions, disclaimers).  No excerpt is needed for this label.
3. **For each label, provide a short rationale explaining your decision.**  The rationale should be separate from the excerpt.
4. **Be very strict with your `supported` and `contradictory` decisions.** Unless you can find straightforward, indisputable evidence excerpts *in the context* that a sentence is `supported` or `contradictory`, consider it `unsupported`. You should not employ world knowledge unless it is truly trivial.

**Input Format:**

The input will consist of two parts, clearly separated:

* **Context:**  The textual context used to generate the response.
* **Response:** The model-generated response to be analyzed.
\end{lstlisting}
\begin{lstlisting}
**Output Format:**

For each sentence in the response, output a JSON object with the following fields:

* `"sentence"`: The sentence being analyzed.
* `"label"`: One of `supported`, `unsupported`, `contradictory`, or `no_rad`.
* `"rationale"`: A brief explanation for the assigned label.
* `"excerpt"`:  A relevant excerpt from the context. Only required for `supported` and `contradictory` labels.

Output each JSON object on a new line.

**Example:**

**Input:**

```
Context: Apples are red fruits. Bananas are yellow fruits.

Response: Apples are red. Bananas are green. Bananas are cheaper than apples. Enjoy your fruit!
```

**Output:**

{"sentence": "Apples are red.", "label": "supported", "rationale": "The context explicitly states that apples are red.", "excerpt": "Apples are red fruits."}
{"sentence": "Bananas are green.", "label": "contradictory", "rationale": "The context states that bananas are yellow, not green.", "excerpt": "Bananas are yellow fruits."}
{"sentence": "Bananas are cheaper than apples.", "label": "unsupported", "rationale": "The context does not mention the price of bananas or apples.", "excerpt": null}
{"sentence": "Enjoy your fruit!", "label": "no_rad", "rationale": "This is a general expression and does not require factual attribution.", "excerpt": null}

**Now, please analyze the following context and response:**

**User Query:**
{user_query}

**Context:**
{context}

**Response:**
{response}
\end{lstlisting}

\noindent\textit{JSON w. double-check} (this template uses the \textit{JSON} template initially, while using the below template to double-check each sentence-level classification) 

\begin{lstlisting}
Your task is to verify whether a given sentence is entailed by a given context or not. Answer only in YES or NO without any additional text. Do not try to avoid answering, or apologize, or give any answer that isn't simply YES or NO.

**Sentence**
{json_dict["sentence"]}

**Context**
{json_dict["excerpt"]}
\end{lstlisting}

\noindent\textit{Span-level} 

\begin{lstlisting}
Your task is to check if a specific Span is accurate to the Evidence.
Generate 'Accurate' if the Span is accurate when verified according to the Evidence or when there is nothing to verify in the Span.
Generate 'Inaccurate' if the Span is inaccurate (contradicts the evidence), or cannot be verified.

**Query**:\n\n{user_query}\n\n**End of Query**\n
**Evidence**\n\n{context}\n\n**End of Evidence**\n
**Response**:\n\n{response}\n\n**End of Response**\n

You are currently verifying **Span {ix+1}** from the Response.
**Span {ix+1}**:\n\n{span}\n\n**End of Span {ix+1}**\n

Is Span {ix+1} accurate or inaccurate when verified according to the Evidence? Point to where in the evidence justifies your answer.
\end{lstlisting}

\noindent\textit{Implicit span-level} 

\begin{lstlisting}
Your task is to check if the Response is accurate to the Evidence.
Generate 'Accurate' if the Response is accurate when verified according to the Evidence, or 'Inaccurate' if the Response is inaccurate (contradicts the evidence) or cannot be verified.

**Query**:\n\n{user_query}\n\n**End of Query**\n
**Evidence**\n\n{context}\n\n**End of Evidence**\n
**Response**:\n\n{response}\n\n**End of Response**\n

Break down the Response into sentences and classify each one separately, then give the final answer: If even one of the sentences is inaccurate, then the Response is inaccurate.

For example, your output should be of this format:
Sentence 1: <Sentence 1>
Sentence 1 label: Accurate/Inaccurate (choose 1)
Sentence 2: <Sentence 2>
Sentence 2 label: Accurate/Inaccurate (choose 1)
Sentence 3: <Sentence 3>
Sentence 3 label: Accurate/Inaccurate (choose 1)
[...]
Final Answer: Accurate/Inaccurate (choose 1)
\end{lstlisting}

\noindent\textbf{Ineligible responses filter.}

The following template is used for both the \textit{user-request-only} and the \textit{user-request-with-context-document} methods. The only difference is whether only the user request or the full prompt are inserted respectively in place of the \lstinline{user_request_or_full_prompt} placeholder. Responses are ineligible if they contain "Major Issue(s)" in instruction following according to all LLM judges.

\begin{lstlisting}
# Rubrics
Your mission is to judge the response from an AI model, the *test* response, calibrating your judgement using a *baseline* response.
Please use the following rubric criteria to judge the responses:

<START OF RUBRICS>
Your task is to analyze the test response based on the criterion of "Instruction Following". Start your analysis with "Analysis".

**Instruction Following**
Please first list the instructions in the user query.
\end{lstlisting}
\begin{lstlisting}
In general, an instruction is VERY important if it is specifically asked for in the prompt and deviates from the norm. Please highlight such specific keywords.
You should also derive the task type from the user query and include the task-specific implied instructions.
Sometimes, no instruction is available in the user query.
It is your job to infer if the instruction is to autocomplete the user query or is asking the LLM for follow-ups.
After listing the instructions, you should rank them in order of importance.
After that, INDEPENDENTLY check if the test response and the baseline response meet each of the instructions.
You should itemize, for each instruction, whether the response meets, partially meets, or does not meet the requirement, using reasoning.
You should start reasoning first before reaching a conclusion about whether the response satisfies the requirement.
Citing examples while reasoning is preferred.

Reflect on your answer and consider the possibility that you are wrong.
If you are wrong, explain clearly what needs to be clarified, improved, or changed in the rubric criteria and guidelines.

In the end, express your final verdict as one of the following three json objects:

```json
{{
  "Instruction Following": "No Issues"
}}
```

```json
{{
  "Instruction Following": "Minor Issue(s)"
}}
```

```json
{{
  "Instruction Following": "Major Issue(s)"
}}
```

<END OF RUBRICS>

# Your task
## User query
<|begin_of_query|>
{user_request_or_full_prompt}
<|end_of_query|>

## Test Response:
<|begin_of_test_response|>
{test_response}
<|end_of_test_response|>

## Baseline Response:
<|begin_of_baseline_response|>
{baseline_response}
<|end_of_baseline_response|>

Please write your analysis and final verdict for the test response. 
\end{lstlisting}